\pdfoutput=1

\documentclass[11pt,a4paper]{article}
\PassOptionsToPackage{svgnames}{xcolor}

\usepackage{times,latexsym}
\usepackage{url}
\usepackage[T1]{fontenc}

\usepackage[acceptedWithA]{tacl2021v1}

\usepackage{xspace,mfirstuc,tabulary}

\newif\iftaclinstructions
\taclinstructionsfalse 
\iftaclinstructions
\renewcommand{\confidential}{}
\renewcommand{\anonsubtext}{(No author info supplied here, for consistency with
TACL-submission anonymization requirements)}

\fi

\iftaclpubformat 

\else

\fi

\usepackage[utf8]{inputenc}
\usepackage{microtype}
\usepackage{inconsolata}
\usepackage{amssymb}
\usepackage{pifont}
\usepackage{pgfplots} 
\usepackage[table]{xcolor}
\usepackage{mdframed}
\usepackage{hyperref}
\usepackage{linguex}
\usepackage[most]{tcolorbox}
\usepackage{amsmath}
\usepackage{multirow}
\usepackage{longtable}
\usepackage[colorinlistoftodos]{todonotes}

\newcommand{\yes}{\tikz[baseline=-0.5ex]\fill[green!55!black] circle (4.5pt);}
\newcommand{\no}{\tikz[baseline=-0.5ex]\fill[red!65!black] circle (4.5pt);}
\newcommand{\opt}{\tikz[baseline=-0.5ex]\draw[gray,thick] circle (4.5pt);}
\usepackage{booktabs}
\usepackage[edges]{forest}

\newcommand{\changed}[1]{\textcolor{black}{#1}}

\definecolor{tagfairness}{RGB}{252,228,236}
\definecolor{tagfairnesstxt}{RGB}{160,30,50}
\definecolor{tagharm}{RGB}{252,228,236}
\definecolor{tagharmontxt}{RGB}{160,30,50}
\definecolor{tagequality}{RGB}{232,245,233}
\definecolor{tagequalitytxt}{RGB}{30,110,55}
\definecolor{tagideology}{RGB}{227,242,253}
\definecolor{tagideologytxt}{RGB}{21,95,180}
\definecolor{tagpro}{RGB}{220,245,240}
\definecolor{tagprotxt}{RGB}{10,110,90}
\definecolor{tagagainst}{RGB}{252,228,236}
\definecolor{tagagainsttxt}{RGB}{160,30,50}
\definecolor{tagneg}{RGB}{255,232,232}
\definecolor{tagnegtxt}{RGB}{170,25,25}
\definecolor{tagemotions}{RGB}{255,248,210}
\definecolor{tagemotionstxt}{RGB}{140,95,0}
\definecolor{tagmorality}{RGB}{238,237,253}
\definecolor{tagmoralitytxt}{RGB}{60,52,200}
\definecolor{tagcrime}{RGB}{238,237,253}
\definecolor{tagcrimetxt}{RGB}{60,52,200}
\definecolor{tagsemframe}{RGB}{242,235,224}
\definecolor{tagsemframetxt}{RGB}{110,55,20}
\definecolor{quotebg}{RGB}{245,245,245}
\definecolor{catcol}{RGB}{100,100,100}

\usepackage{tabularx}
\usepackage{colortbl}
\usepackage{array}
\newcommand{\atag}[3]{%
  \setlength{\fboxsep}{1.5pt}%
  \colorbox{#1}{\textcolor{#2}{\scriptsize\textbf{#3}}}%
}
\newcommand{\lbl}[1]{{\scriptsize\textcolor{catcol}{\textsc{#1}}}}

\usepackage{pgf} 
\usepackage[table]{xcolor}
\definecolor{sdmin}{RGB}{180,230,180}  
\definecolor{sdmax}{RGB}{240,160,150} 

\usepackage{ifthen}
\definecolor{sdc0}{RGB}{180,230,180}
\definecolor{sdc1}{RGB}{200,228,168}
\definecolor{sdc2}{RGB}{220,224,155}
\definecolor{sdc3}{RGB}{232,218,145}
\definecolor{sdc4}{RGB}{242,205,135}
\definecolor{sdc5}{RGB}{244,185,128}
\definecolor{sdc6}{RGB}{242,168,130}
\definecolor{sdc7}{RGB}{240,148,128}

\newcommand{\scorecell}[2]{%
  \ifthenelse{\equal{#2}{0.00}}{\cellcolor{sdc0}}{}%
  \ifthenelse{\equal{#2}{0.58}}{\cellcolor{sdc1}}{}%
  \ifthenelse{\equal{#2}{1.00}}{\cellcolor{sdc2}}{}%
  \ifthenelse{\equal{#2}{1.15}}{\cellcolor{sdc3}}{}%
  \ifthenelse{\equal{#2}{1.53}}{\cellcolor{sdc4}}{}%
  \ifthenelse{\equal{#2}{1.73}}{\cellcolor{sdc5}}{}%
  \ifthenelse{\equal{#2}{2.00}}{\cellcolor{sdc6}}{}%
  \ifthenelse{\equal{#2}{2.31}}{\cellcolor{sdc7}}{}%
  #1 $_{{\pm}#2}$%
}

\title{Structuring the Space of \textit{Perspectives}}

\author{
  Agnese Daffara
  \\
  University of Stuttgart
  \\
  \footnotesize\texttt{agnese.daffara@ims.uni-stuttgart.de}
  \And
  Sebastian Pad\'o
  \\
  University of Stuttgart
  \\
  \footnotesize\texttt{pado@ims.uni-stuttgart.de}
  \And
  Tanise Ceron
  \\
  Bocconi University
  \\
  \footnotesize\texttt{tanise.ceron@unibocconi.it}
}

\date{}

\begin{document}
\maketitle
\begin{abstract}

The same event can be reported 
from different \textit{perspectives} depending on the experiences, background, and beliefs of the writer or speaker. 
A variety of NLP areas engage with  perspectives, spanning from text analysis to algorithm optimization. 
A wide range of operative concepts (such as \textit{stances}, \textit{sentiment}, \textit{frames}, and \textit{arguments}) has been used to capture perspectives in texts, however the precise relationships among those concepts remain unclear. Arguably, a deeper theoretical understanding of these concepts would empower more effective research on perspectives. 
In this paper, we address \changed{this gap} by  reviewing the  space of perspectives in NLP \changed{and defining} a set of properties that help distinguishing perspective-related concepts. Our analysis leads us to posit a hierarchy which organizes these concepts linearly along a single axis. Finally, we show how this principled conceptual hierarchy can help researchers navigate the field and select operationalizations of perspective that align with their specific research objectives.
\end{abstract}

\section{Introduction}
\label{sec:intro}

\begin{figure}[tb!]
  \centering
  \includegraphics[width=1\linewidth]{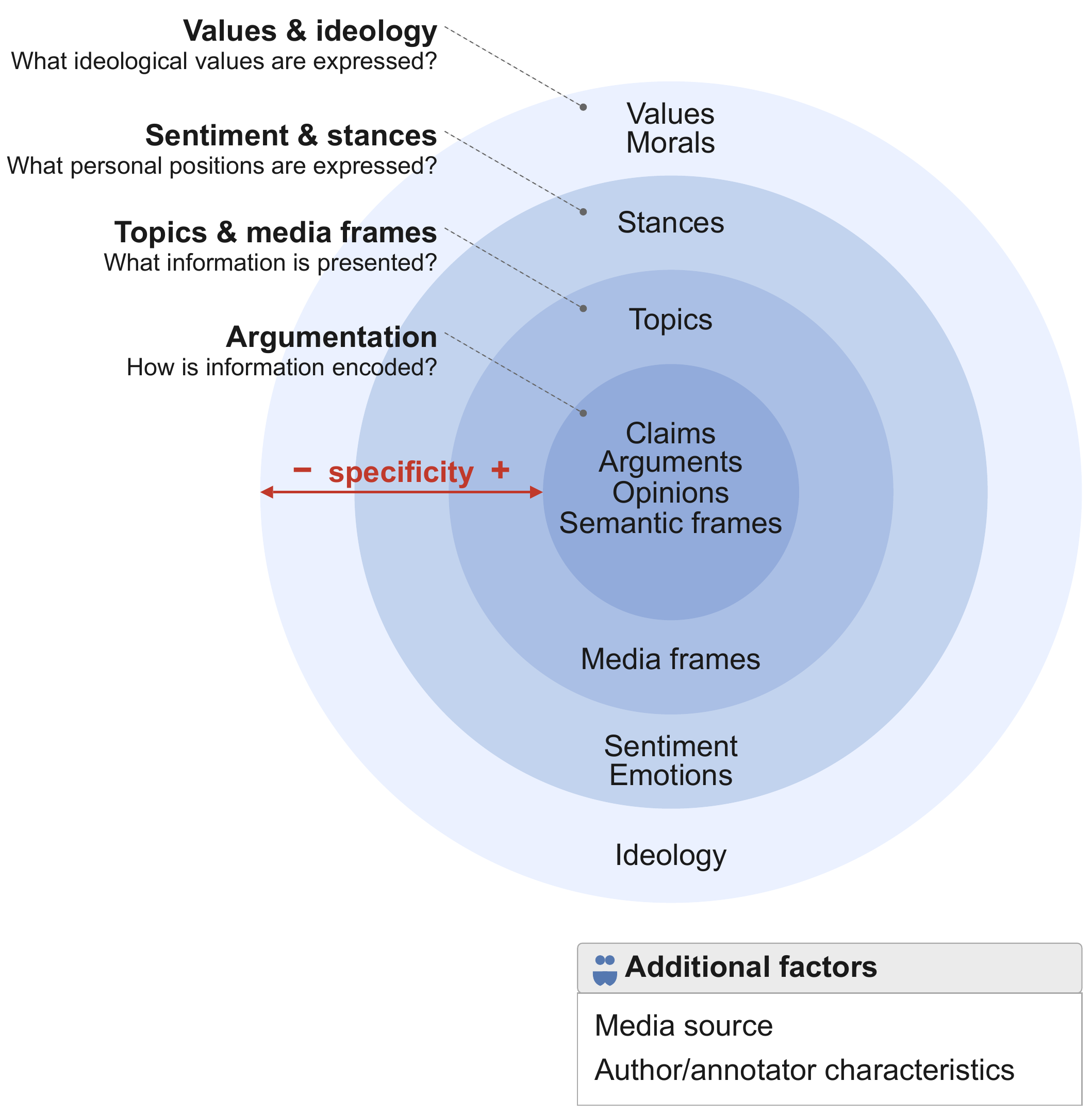}
  \caption{A model of perspective with key concepts related to \textbf{perspective}. The scheme arranges them on a \textit{specificity} axis, ranging from ideological concepts to specific linguistic devices, as described in §\ref{sec:hierarchy}. 
  Extra-textual factors are shown in a separate box.}
  \label{fig:typo}
\end{figure}

Every day we are exposed to a wide variety of information through different text types, from personal-related content like product reviews and social media posts, to official productions like political debates and news articles. While some texts explicitly express personal opinions, others aim to report factual information. Yet, all of them project some \textbf{perspective} onto the world. 
Indeed, perspective-taking is a prerequisite of human communication \cite{graumann2008perspective}. 
The NLP community has long studied perspectives, but there is a growing interest in recent years\footnote{The number of ACL papers with \textit{perspective(s)} in abstract or title grew from 17 in 2015 to 150 in 2020 and 624 in 2025.}. 

This trend has unfolded across many (sub)communities and under a variety of labels (cf. Figure \ref{fig:typo}). 
The diversity of concepts and frameworks makes it challenging to clarify the connections between different studies, to identify gaps in existing research, and to uncover the theoretical backgrounds and assumptions underlying individual studies. Imagine, for example, that a researcher aims to build a multi-perspective news recommender or to diversity opinions in language models' generated text: what concepts and what research should they be aware of? As \citet{reuver-etal-2021-nlp} note, bridging perspective frameworks from the domain of Natural Language Processing (NLP) is essential to advance democratic values in information consumption and dissemination.

\paragraph{The Term \textit{Perspective}}
\label{sec:term}

A \textit{perspective} generally indicates a viewpoint on reality. Historically, the term has been used in multiple ways across research fields. In narrative theory, it simply refers to the positioning of a viewpoint within a particular character or narrator in the discourse \citep{sanders1993linguistic}. \changed{From a cognitive viewpoint, it has been demonstrated that reality itself is a collection of perspectives \cite{basile2022bias}. }
In discourse studies and linguistics, it has been widely studied how semantic and syntactic choices trigger \textit{perspectivization} (or \textit{vantage point taking}), by representing a state of affair in terms of actor roles and their viewpoints; e.g., whether the perpetrator of a murder is presented as the responsible agent \citep{graumann2008perspective}. Recently, the research program of \textit{perspectivism} adopted the term to denote the entirety of socio-demographic, cultural, and individual traits in annotations and modeling \cite{frenda2024perspectivist}. 
\changed{In this paper, we address perspectives directly embedded in the text and discuss extra-textual traits as additional factors (cf. Figure \ref{fig:typo})}. Building on its generic usage in NLP literature, we adopt \textit{perspective} as an umbrella term, comprising related concepts in a way to be further clarified.

\paragraph{Relevance in NLP} The interest in identifying perspectives in texts 
has mainly two aims: i) perspective recognition; 
ii) perspective generation. 
Regarding the first aim, \textit{perspective} traditionally denotes a political orientation \citep{pang2008opinion, kuccuk2020stance}; e.g., \textsc{liberal} vs. \textsc{conservative}. However, it is commonly expanded in its meaning to indicate other related notions, like \textit{opinions} \cite{morante2020annotating}, \textit{stances} \cite{klebanov2010vocabulary, roy2023tale}, \textit{frames} \cite{liu-etal-2019-detecting, alashri2015animates}, \textit{sentiment} \cite{yu2003towards, greene2009more}, and \textit{claims} \cite{chen2019seeing}. 

Concerning the second aim, the rise of LLMs has introduced new research questions about perspectives: what kind of data and annotations are used for training? Are they biased toward specific viewpoints \cite{lin2025investigating,ceron2025political}? Can these models preserve multiple and minor perspectives in summarization \cite{van-der-meer-2024-facilitating} and reproduce different opinions in generation \cite{hu-etal-2025-debate}? Generating diverse perspectives in AI applications is desirable to promote alignment with democratic values and ensure that certain opinions are not underrepresented \cite{sorensen-2024-pluralistic}. 

\paragraph{Previous Work}
A number of surveys focus on perspective-related concepts and tasks. This enables a detailed investigation, but limits the possibility of connecting and organizing them -- the gap we address in this paper. Among
others, \citet{pang2008opinion} provide an overview on
opinion mining and sentiment analysis, \citet{munezero2014they} on subjectivity-related terms; \citet{kuccuk2020stance} on stance detection, \citet{doan2022survey} on political perspective detection,
\citet{otmakhova-etal-2024-media} on media frames, and
\citet{rodrigo2024systematic} on media bias (cf. Appendix \ref{app:papers} for more references). 

\changed{All these concepts relate to how a viewpoint is
expressed in language and are thus characterized by some shared linguistic
features. A line of work in linguistics \cite{hunston2000evaluation, benamara2017evaluative} uses the term
\textit{evaluation} to denote ``the speaker or writer's attitude or
stance towards, viewpoint on, or feelings about the entities or
propositions that she or he is talking about'' (\citealt{hunston2000evaluation}, p. 5). This clarifies the linguistic surface of perspective and roots the discussion in the domain of \textit{evaluation}, as opposed to \textit{factual knowledge}.}

\paragraph{Contributions and Structure}

Our paper aims at improving the state of the art in perspective-related research by 
asking two research questions:


\begin{itemize}
    \item \textbf{RQ1}: What concepts are used to identify perspectives in text?
    \item \textbf{RQ2}: How are these concepts related?
\end{itemize}
%
To answer these questions, we first introduce and characterize the space of perspective-related concepts based on the literature 
(§\ref{sec:perspectives}). We then organize and structure the conceptual space with a property annotation\footnote{This type of analysis is also referred to as \textit{content analysis} in social sciences, but we keep NLP terminology.}, clustering analysis and Principal Component Analyis (PCA) (§\ref{sec:conceptualization}). The outcome \changed{of the analysis is the \textit{model of perspective} shown in Figure~\ref{fig:typo}: we find that clusters of perspective-related concepts form a hierarchy along a linear scale of conceptual and linguistic \textit{specificity}.} \changed{To make this insights actionable, we propose a decision tree (Figure~\ref{fig:diagram}) to help researchers make informed choices among the discussed concepts that match their specific research goals (§\ref{sec:discussion}).}



\section{Paper Collection}
\label{sec:collection}
\changed{Our paper is not a full survey, but rather a conceptual organization, supported by a synthesized literature review with the goal of capturing the space of  perspective-related concepts together with their definitions and characteristic properties. To understand the concepts related to perspectives in NLP, we first conduct a regular expression based search on the ACL Anthology bibliography and select 60 papers (see Appendix~\ref{app:acl_search} for details).}

\changed{Since this search did not capture some foundational work that contributed to the conceptualization of perspective in NLP, neither the literature from neighbor areas (e.g, communication science), we continue collecting papers manually on Google Scholar, by following citation trails in references and related works. Papers are included in the analysis if they: i) provide a definition of \textit{perspective}; ii) use \textit{perspective} in a conceptual way that distinguishes it from similar work; iii) establish a conceptual grounding for a certain term. After some iterations, we organize all papers in a bottom-up fashion, which resulted in a final set of relevant concepts. The final number of collected papers is 227 (full list in Appendix \ref{app:papers}).}
\newline
\section{Perspective-related Concepts}
\label{sec:perspectives}


Based on our analysis of the extant literature, we collect a total of 15 concepts relevant for \textit{perspectives}. 
For each concept, we characterize it and briefly discuss relevant textual features and methods used in computational modeling ("Perspective Signals"). \changed{Discriminative properties of each concept are marked in \textbf{bold};
linguistic levels are indicated in \textit{italics}.} Section \ref{sec:interactions} discusses how these concepts relate to each other \changed{and provides two unified examples}. Further definitions and additional examples can be found in Appendix \ref{app:codebook}.

\subsection{Morals and Values}
\label{sec:values}
\changed{We include morals and values because they are increasingly
studied in NLP as the foundation of perspective given their proximity with ideological positions \cite{graham2009liberals}. Values motivate arguments by influencing the positions adopted and the justifications offered for them \cite{kiesel-etal-2022-identifying, atkinson2021value, van-der-meer-2024-facilitating}. They form ``the basis for all processes of
evaluation'' \cite{van1998ideology}, being "the intrinsic goods or ideals that individuals pursue or cherish" \cite{sorensen2024value},
such as \textsc{Freedom} and \textsc{Equality}. 
Morals are frequently studied together in the NLP literature, but they origin from distinct frameworks \cite{d_f_2026}; they are \textbf{societal} and encode a \textbf{shared} judgment
of what is \textit{right} or \textit{wrong} \cite{vida-etal-2023-values, graham2009liberals, entman1993framing},
e.g., the moral principle that \textit{causing harm is wrong} is accepted across cultures.}

\paragraph{Perspective Signals}

\changed{Values are grounded in psychological frameworks (e.g., \citet{schwartz1992universals}) and are often operationalized through social surveys, such as the
World Values Survey \cite{inglehart2000world}. While values are difficult to spot in text because they are high-level constructs, some evaluation marks can be recognized, such as the mention of goals or (non-)achievements \cite{hunston2000evaluation}. Morals are often studied with reference to the Moral Foundations Theory (MFT) \cite{graham2009liberals}, which identifies five virtue/vice dimensions: \textsc{Care/Harm},
\textsc{Fairness/Cheating}, \textsc{Loyalty/Betrayal},
\textsc{Authority/Subversion}, and \textsc{Purity/Degradation}. The Moral Foundations Dictionary \cite{graham2009liberals} maps \textit{lexical} items to these dimensions. Morals and values are
also operationalized as frames -- some values, such as \textsc{security}, overlap with Media Frames labels; at the \textit{semantic} level, \textit{morality frames} associate the MFT dimensions with agents and objects
\cite{roy2021identifying}. Recent work on morals and values in NLP focuses mainly on generating \textit{pluralistic values}
\cite{sorensen2024value} and assessing LLMs' ideological
\cite{ceron-et-al-2024, benkler2023assessing} and moral \cite{abdulhai2024moral}
alignment or inducing it through reinforcement learning.}



\subsection{Ideology}
\label{sec:ideology}

\changed{While morals and values provide the overarching beliefs guiding decisions, ideology is the coherent system that organizes them
into a stable set of ideas shared by a group.} \changed{Ideology is \textbf{stable} because it is anchored in a
shared system of core values that
function as evaluative criteria across topics and contexts; just like the grammar of a language, which is the reference point for its rules (\citealt{van1998ideology}, p. 56). Because it is anchored in these group-level
commitments, ideology operates
at a higher level of \textbf{abstraction} than stance or sentiment (which are granular and variable),
and can be interpreted as a cluster of aligned opinions
\cite{van1998ideology, doan2022survey}. In NLP, ideology is used in \textbf{political} domains \cite{pang2008opinion}.} It manifests as (i) a
position in a debate (e.g.\ \textsc{pro-Israel} vs.\
\textsc{pro-Palestine}), (ii) a political leaning on a spectrum
(e.g.\ \textsc{Left} vs.\ \textsc{Right}), or (iii) a party affiliation
(e.g.\ \textsc{Democratics} vs.\ \textsc{Republicans}).

The task of \textit{ideology bias detection} classifies texts as
\textsc{biased} or \textsc{unbiased} (or on a scale in-between),
measuring the degree of ideological skew \cite{rodrigo2024systematic}, which arguably assumes the existence of neutral, factual texts
\cite{vargas2023predicting}. 
Note that while \textit{bias detection} identifies whether a text is
ideologically skewed without specifying orientation, \textit{ideology
detection} assigns it to a specific group. 
Early work framed the task as \textbf{binary} classification between two opposing
views, such as the Israeli--Palestinian conflict \citep{lin2006side}. 
Political leaning and party detection are instead typically modelled as
\textbf{multi-class}, regression or scaling problems, situating texts along a political
scale \cite{budak2016fair, kiesel2019semeval, ceron2022optimizing}, for
example from \textsc{conservative} to \textsc{liberal}.

\paragraph{Perspective Signals}
\changed{At the \textit{lexical} level, the most informative signals are
one-sided terms and sticky bigrams \cite{klebanov2010vocabulary,
recasens2013linguistic, monroe2008fightin}, for example,
\textit{illegal aliens} is linked to conservative discourse
\cite{webson-etal-2020-undocumented}. Early approaches exploited
these via bag-of-words \cite{lin2006side, laver2003extracting,
slapin2008scaling} and n-grams \cite{hardisty2010modeling}; because
such patterns overlap with topic distributions, perspectives have
also been modelled via LDA (§\ref{sec:topics}), though this risks
reducing ideology to surface frequencies.
At the \textit{semantic} and \textit{syntactic} levels, factive
verbs, lexical entailments, hedges \cite{greene2009more,
recasens2013linguistic}, and constructions like the passive voice
convey ideological positioning without overt evaluation. At the
\textit{pragmatic} level, metaphor \cite{sengupta-etal-2024-analyzing}
and rhetorical strategies \cite{huguet-cabot-etal-2020-pragmatics}
carry meaning beyond literal content (cf. Table~\ref{tab:signals}).}

Neural approaches capture signals across all levels through
word embeddings \cite{iyyer2014political, gangula2019detecting,
li-goldwasser-2019-encoding, alzhrani2022political}, at the cost of
interpretability \cite{martinez-etal-2024-balancing}. Recent work has
leveraged encoder-based LLMs \cite{baly-etal-2020-detect} and decoder-based ones \cite{kim2023multi, da2023overview}, enriched with social network relations \cite{li-goldwasser-2019-encoding, baly-etal-2020-detect}, Wikipedia \cite{li-goldwasser-2021-mean, feng2021kgap}, or political speeches \cite{jakob-etal-2024-augmented}. Hybrid methods combine text with knowledge graphs
\cite{zhang-etal-2022-kcd}.

\begin{table}[tb!]
\small\centering
\setlength{\tabcolsep}{5pt}
\begin{tabular}{p{3.5cm}p{3.5cm}}
\toprule
\textbf{Phenomenon} & \textbf{Example} \\
\midrule
\multicolumn{2}{l}{\cellcolor{gray!15}\textit{Lexical}} \\
One-sided terms       & \textit{pro-life} \\
Sticky bigrams        & \textit{illegal aliens} \\
Evaluative language   & \textit{suitable} \\
Subj.\ intensifiers   & \textit{fantastic} \\[3pt]
\multicolumn{2}{l}{\cellcolor{gray!15}\textit{Semantic}} \\
Factive verbs         & \textit{reveal} \\
Assertive verbs       & \textit{say} \\
Lexical entailments   & \textit{murdered} \\
Hedges                & \textit{possibly} \\[3pt]
\multicolumn{2}{l}{\cellcolor{gray!15}\textit{Syntactic}} \\
Passive constructions & \textit{mistakes were made} \\[3pt]
\multicolumn{2}{l}{\cellcolor{gray!15}\textit{Pragmatic}} \\
Metaphor              & \textit{real beef} \\
Implicature           & [implicit blame] \\
\bottomrule
\end{tabular}
\caption{Signals of ideological bias \changed{by level of linguistic
analysis }\cite{yano2010shedding, recasens2013linguistic,
greene2009more, sengupta-etal-2024-analyzing}.}
\label{tab:signals}
\end{table}


\subsection{Stances}
\label{sec:stance}
Stance indicates an ideological position towards a target. \changed{It is usually treated together with sentiment as an evaluative and \textbf{affective} concept. However, it may also be \textit{epistemic} if there is no affective component \cite{kiesling2018interactional, kuccuk2020stance}}. 
Typical labels are \changed{\textsc{Against}/\textsc{Negative}, \textsc{Neutral}/\textsc{Neither}, and \textsc{Pro}/\textsc{Favor}/\textsc{Positive}}. While \textit{binary ideology detection} is sometimes referred to as \textit{stance detection}, there are important distinctions between the two: (i) ideology \changed{is a stable overarching system, whereas stance is individually \textbf{variable} (cf. §\ref{sec:ideology})}, and (ii) ideology is target-generic, whereas stance is \textbf{target-specific}, requiring one \changed{or more} explicit or inferable targets \cite{kuccuk2020stance}\changed{:} entities (e.g. Donald Trump), policy issues (e.g. border control), events (e.g. the approval of a new policy), or claims (e.g. "we need to increase border control to ensure national security"). Stance also overlaps with \textit{entity-based} or \textit{aspect-based sentiment analysis}, which seeks to identify emotional attitude toward a target (§\ref{sec:sentiment}); however, stance reflects evaluative alignment rather than \changed{emotional} tone. As  \citet{hasan2012predicting} note, the same document may have negative sentiment expressions but a positive stance (cf. Figure \ref{fig:examples}).

\paragraph{Perspective Signals}
Stance is typically conveyed through the structure of argumentation,
which makes it difficult to capture for simple bag-of-words approaches. Pioneer stance detection tasks  integrated
\changed{\textit{lexical}} subjectivity and polarity features with \changed{parse trees and {\textit{discourse}}-level}
argumentative features \cite{wiebe2005annotating,
somasundaran-wiebe-2010-recognizing, hasan2012predicting, bar2017stance,
anand2011cats}. These features remain latent when using neural networks
\cite{roy2023tale, mohammad2016semeval}. The detection can be enriched
by integrating \changed{\textit{semantic} information} like entities, their roles,
and associated sentiment \cite{roy2023tale}.

\subsection{\changed{Sentiment and Emotions}}
\label{sec:sentiment}
We have seen that ideology reflects stable, overarching belief systems
shared by a group (§\ref{sec:ideology}) and stance captures the
author's ideological position towards a target (§\ref{sec:stance}). \changed{In contrast, sentiment and opinions indicate a \textbf{subjective} response with an \textbf{affective} component.} In fact,
they are traditionally linked to the area of subjectivity detection,
where they were originally theorized as \textit{private states}, i.e., mental states that are not accessible to objective observation or
verification \cite{Quirk1985G}.

Sentiment can be interpreted in two ways: (i) in a general sense, it
refers to instances of perspective expressed in a text, which is why
\textit{sentiment analysis} and \textit{opinion mining} are traditionally
treated as equivalent tasks \cite{pang2008opinion}; (ii) in a more
fine-grained view, it reflects the \textbf{polar} orientation
(\textsc{positive}, \textsc{neutral}, \textsc{negative}) of an opinion,
whereas an opinion represents the full perspective expression
\cite{munezero2014they}. For example, the sentence ``illegal aliens are
ruining the country'' is a fully opinion expression including a
\textsc{negative} sentiment. Sentiment can also be referred to specific
targets or entities (\textit{entity-based}) or to specific attributes of the target (\textit{aspect-based}) \changed{\cite{ronningstad2024entity, kuccuk2020stance}}.

\changed{Emotions are \textbf{affective} states that go
\textbf{beyond polar} orientation to specify the type of emotional response (such as \textsc{fear}, \textsc{anger}, or \textsc{sadness}) typically grounded in psychological models such as Ekman's basic
emotions \cite{ekman1992argument} or Plutchik's wheel
\cite{plutchik1980general,plaza-del-arco-etal-2024-emotion}.}

\paragraph{Perspective Signals}
Early approaches to sentiment analysis \changed{and emotion detection} used
subjectivity features as a proxy. \changed{At the
\textit{lexical} level, they relied on static methods:} unigrams
\cite{pang2002thumbs}, pre-compiled subjectivity lexicons
\cite{liu2005opinion, yu2003towards, wilson2005recognizing, mohammad2013nrc} and bootstrapped patterns
\cite{riloff2003learning}\changed{. At the \textit{semantic} and \textit{discourse} levels, dynamic methods exploit} contextual meaning
\cite{wilson2005opinionfinder}, WordNet relations
\cite{choi2014+}, constituents \cite{kim2004determining}, and
 dependency patterns combined with discourse-level cues
\cite{hasan2012predicting}.
A key resource supporting this line of research is the MPQA (Multi-Perspective Question Answering) corpus \cite{wiebe2005annotating}, comprising news articles annotated with subjectivity, entity- and event-level sentiment \cite{deng2015mpqa}. 

\subsection{Opinions}
\label{sec:opinion}

\citet{munezero2014they} provide an overview of the definitions and representations of \textit{opinion}. First, there is a shared intuition that opinion involves some degree of uncertainty, as it is not factual but instead tied to personal beliefs, just like sentiment. Second, opinion is a \textbf{structured} construct that can be decomposed into various \textbf{sub-components} (see below). Note that also sentiment can be structured into event-level components; the key distinction between these two concepts seems to lay in the facts that (i) an opinion can lack a sentiment, like in the sentences “Bin Laden is hiding in Pakistan” or "I believe the word is flat" \cite{kim2004determining} and (ii) an opinion can be identified with the \textbf{linguistic expression} itself (e.g., "Mary said the dress is beautiful")\footnote{In our conceptual analysis in §\ref{sec:conceptualization}, we adopt this interpretation of opinion as a concrete perspective expression, and consider it a discourse-level concept.}, while sentiment tends to denote an abstract attitude mapped onto polar labels (e.g. \textsc{positive}). 

\paragraph{Perspective Signals}

Opinion mining often employs a structured conceptualization
at the \changed{\textit{semantic} level.} According to
\citet{kim2004determining}, an opinion consists of four elements: (i)
the topic, (ii) the holder, (iii) the claim, and (iv) optionally, the
sentiment. Other proposed components include the
features of the target (aspects) and the time when the opinion is
expressed \cite{liu2010sentiment}.
These conceptualizations are \changed{extended by}
\citet{van2016grasp}, one of the few proposals for a
structured representation of perspective. They proposed an annotation scheme with: (i) the event structure, (ii) the attribution (relationship between the source and the target), (iii) the factuality
(certainty, polarity and time), and (iv) the opinion (sentiment).
The framework was later used to annotate a corpus of news items about
Covid-19 \cite{morante2020annotating}, where perspectives were defined as
``relations between the source of a statement (i.e., the author or
another entity [...]) and a target in that statement
(i.e., an entity, event, or (micro-)proposition)''. In this sense,
structured opinions were seen as perspective expressions operating at
the \changed{\textit{semantic-pragmatic} level}, with sentiment as an ideological
sub-component.
%
\changed{Table \ref{tab:properties_sent} summarizes  discriminative properties of opinion-related concepts.}

\begin{table}[tb!]
\centering\small
\setlength{\tabcolsep}{5.5pt}
\begin{tabular}{p{1.4cm}p{0.7cm}p{0.7cm}p{0.7cm}p{0.8cm}p{1cm}}
\toprule
 & \textit{Sub-comp.} & \textit{Polar} & \textit{Target} & \textit{Stable} & \textit{Affective} \\
\midrule
Ideology  & \no  & \opt & \opt & \yes & \no  \\
Stance    & \opt & \yes & \yes & \no  & \opt  \\
Sentiment & \opt & \yes & \opt & \no  & \yes \\ {Emotions} & {\opt} & {\no} & {\opt} & {\no} & {\yes} \\
Opinion   & \yes & \no  & \yes & \no  & \opt \\
\bottomrule
\end{tabular}
\caption{Discriminative properties of \changed{opinion-related} concepts
describing whether they have sub-components, are polar, have a target,
are stable, and have an affective tone.
\yes~yes; \opt~optional; \no~no. \changed{All properties are discussed in the
literature review (marked in \textbf{bold}).}}
\label{tab:properties_sent}
\end{table}

\subsection{Claims and Arguments}
\label{sec:claims}

Argumentation lies at the core of perspective and can be understood as a basis for representing it \cite{van-der-meer-2024-facilitating}. Therefore, we include in our review the notions of \textit{claim} (a statement functioning as the \textbf{minimal} unit of argumentation) and \textit{argument} (a \textbf{set} of statements composed of premises and conclusions) \citep{govier2010practical}. While opinions describe \textit{what} people think about a topic or product (cf. §\ref{sec:opinion}), arguments explain \textit{why} they hold these opinions \citep{lauscher2022scientia}. Although \textit{opinion mining} and \textit{argument mining} are distinct tasks, their boundaries can be blurry, because opinion mining may also involve identifying argumentative motivations behind a sentiment \cite{cabrio2018five}. Since argumentation provides the foundational layer for perspective detection, in the next paragraph we focus on how it is leveraged for higher-level perspective detection. 

\paragraph{Perspective Signals}

In supervised approaches, argumentation n-grams \changed{involving the \textit{discourse} level} have been shown to be more effective than \textit{lexical} sentiment features for stance detection, as they encode reasoning patterns \changed{\cite{somasundaran-wiebe-2010-recognizing}} (cf. §\ref{sec:stance}). Indeed, the MPQA corpus annotates trigger expressions of positive and negative argumentation (e.g., \textit{be important to, would be better, cannot imagine, we don’t need}). 

In unsupervised approaches, a perspective can be seen as an aggregation of arguments. Clusters of similar arguments (and therefore similar opinions) can reveal overarching belief systems and be interpreted as ideological groups \cite{abu2012subgroup, abu2013identifying, chen2017opinion} (cf. §\ref{sec:ideology}) or can group texts by frame \cite{de2005news, reimers-etal-2019-classification}, where a frame is defined as “a set of arguments that shares an aspect” \cite{ajjour2019modeling} (cf. §\ref{sec:frames}). Only a few contributions propose a more structured approach. \citet{carlebach2020news} use a five-step pipeline for perspective-oriented news aggregation: topic modeling, hypothesis extraction, semantic similarity on hypotheses, premise extraction, and textual entailment. \citet{chen2019seeing} extract distinct arguments associated with a claim, which collectively constitute a spectrum of perspectives (\textit{perspectrum}). Overall, leveraging argumentation structure for perspective detection is still an open research direction \citep{lauscher2022scientia}.

\subsection{Frames}
\label{sec:frames}

According to \citet{boydstun2013identifying}, \textit{framing} means “portraying an issue from one perspective to the necessary exclusion of alternative perspectives”. In some work, \textit{frame} and \textit{perspective} are used as synonyms \cite{alashri2015animates, liu-etal-2019-detecting}. When we frame something, we do three things: (i) selecting: choosing what to present and what not to; (ii) focussing: highlighting or \textbf{emphasizing} some parts;  (iii) embedding: presenting some information as the part and some other as the whole \cite{van2025discourse}. These processes take place at the cognitive level (mental representations of the world), the semantic level (choosing what linguistic structures to use), and the communicative level (impact on the audience) \cite{otmakhova-etal-2024-media}. 
However, defining framing is "notoriously slippery" \cite{boydstun2013identifying, field2018framing} because there are various ways of analyzing it: one can use, for example, \textbf{topic-like} dimensions such as \textit{media frames} \cite{entman1993framing}, \textbf{semantic patterns} such as \textit{semantic frames} \cite{fillmore1976frame} and \textit{connotation frames} \cite{rashkin2016connotation}, narrative dimensions such as \textit{narrative frames} (e.g. \textsc{Hero}, \textsc{Victim}) \cite{otmakhova-etal-2024-media}, or morality dimensions such as \textit{morality frames} (e.g. \textsc{Care/Harm} \cite{roy2021identifying} \changed{(cf. §\ref{sec:values})}. These different ways of detecting framing communicate with each other, for example, the narrative role \textsc{perpetuator} can be associated with the semantic role \textsc{agent}. In this section, we focus on the first two types.

\textit{Media frames} or \textit{communication frames} are framing dimensions found in media. They can be issue-generic or issue-specific and the labels can be defined in an inductive or deductive fashion \cite{de2005news}. A famous annotation framework is the Media Frame Corpus \cite{card2015media}, including 15 labels (e.g., \textsc{Morality}, \textsc{Economic}, \textsc{Health and Safety}), widely adopted in media studies \cite{khanehzar2019modeling, khanehzar2021framing, mendelsohn-etal-2021-modeling, mulder2021operationalizing, gilardi2023chatgpt, piskorski2023semeval}. 
\textit{Semantic frames}, on the other hand, study framing through linguistic structures and semantic roles (e.g. \textsc{killing}: the \textsc{killer} or \textsc{cause} causes the death of a \textsc{victim}). The theory of \textit{semantic frames} was introduced by \citet{fillmore1976frame} and operationalized as FrameNet \cite{baker1998berkeley}. Framing is a  \textit{perspectivization} where cognitive dispositions are induced and perpetuated through language \cite{minnema-etal-2022-sociofillmore}. The main limitation is that semantic frames are primarily used to investigate specific  issues, such as femicides \cite{minnema-etal-2022-dead} and car crashes \cite{te2020framing}, as the specificity of the patterns impedes cross-domain generalization.

\paragraph{Perspective Signals}

Early work in media frame detection relied on topic models, subtracting framing features from topic representations (cf. §\ref{sec:topics}). Traditional supervised classification techniques mainly leveraged n-grams and \changed{\textit{lexical}} features \cite{baumer2015testing}. Following approaches explored embedding-based lexicon expansion \cite{field2018framing}, neural networks \citep{naderi2017classifying, liu-etal-2019-detecting, morstatter2018identifying}, fine-tuning of pre-trained models \cite{liu-etal-2019-detecting, khanehzar2019modeling, mendelsohn-etal-2021-modeling, kwak2020systematic} and prompting \cite{piskorski2023semeval, gilardi2023chatgpt}. 
Entity-level framing has been explored to characterize latent personas \cite{card-etal-2016-analyzing} and analyze the portrayal of social actors \cite{ziems-yang-2021-protect-serve, roy2023tale, masini2017actor}.

Semantic frames are linguistically more interpretable and directly tied to concrete linguistic schemes that involve \changed{\textit{lexical}} units and \changed{\textit{semantic}} roles. These frames can influence the way information is conveyed; for example, in reporting a femicide, the choice of \textit{dead} over \textit{murdered} can shift responsibility onto the victim and obscure agency. A few multilingual detection tools exist in NLP, including LOME \citep{xia-etal-2021-lome} and SocioFillmore \cite{minnema-etal-2022-sociofillmore}.

\begin{figure*}[tb]
\begin{tcolorbox}[
  title=Unified annotated examples,
  breakable=false,
  fonttitle=\small\bfseries,
  colback=white, colframe=gray!40,
  left=3pt, right=3pt, top=2.5pt, bottom=2.5pt,
  toptitle=2pt, bottomtitle=2pt
]
\small\sloppy

\begin{minipage}[t]{0.485\linewidth}
\textbf{Ex.\,1} \citep{lin2006side}

\smallskip
\colorbox{quotebg}{\parbox{\dimexpr\linewidth-2\fboxsep}{\itshape
``The inadvertent killing by Israeli forces of Palestinian civilians ---
usually in the course of shooting at Palestinian terrorists --- is considered
no different at the moral and ethical level than the deliberate targeting of
Israeli civilians by Palestinian suicide bombers.''}}

\smallskip
\lbl{Morals \& Values}
\atag{tagfairness}{tagfairnesstxt}{Fairness} $\cdot$ \atag{tagequality}{tagequalitytxt}{Universalism}\\[2pt]
\lbl{Ideology \& Stance}
\atag{tagpro}{tagprotxt}{Pro-Israeli}\\[2pt]
\lbl{Sentiment \& Emotion}
\atag{tagneg}{tagnegtxt}{Negative} $\cdot$ \atag{tagemotions}{tagemotionstxt}{Indignation}\\[2pt]
\lbl{Topic} Israeli--Palestinian war \;
\lbl{Media Frame}
\atag{tagmorality}{tagmoralitytxt}{Morality}\\[2pt]
\lbl{Semantic Frame}
\atag{tagsemframe}{tagsemframetxt}{Killing} [\textit{inadvertent}; intent absent] vs.\ \atag{tagsemframe}{tagsemframetxt}{Targeting} [\textit{deliberate}; intent foregrounded]\\[2pt]
\lbl{Opinion} holder: author $\cdot$ both actions are morally equivalent\\[2pt]
\lbl{Argument} premise: both sides cause civilian deaths $\cdot$ conclusion: neither is more culpable\\[2pt]
\lbl{Claim} \textit{``Both actions are no different at the moral and ethical level.''} [central statement]
\end{minipage}
\hfill{\color{gray!40}\vrule width 0.4pt}\hfill
\begin{minipage}[t]{0.485\linewidth}
\textbf{Ex.\,2} \citep{hasan2012predicting}

\smallskip
\colorbox{quotebg}{\parbox{\dimexpr\linewidth-2\fboxsep}{\itshape
``Do you really think that criminals won't have access to guns if the federal
government bans guns? A firearm ban will only cause deaths of innocent citizens.''}}

\smallskip
\lbl{Morals \& Values}
\atag{tagharm}{tagharmontxt}{Harm} $\cdot$ \atag{tagequality}{tagequalitytxt}{Security}\\[2pt]
\lbl{Ideology \& Stance}
\atag{tagideology}{tagideologytxt}{Right-leaning} \atag{tagideology}{tagideologytxt}{Conservative} $\cdot$ neg.\ expressions yet \atag{tagpro}{tagprotxt}{pro-gun} alignment \citep{hasan2012predicting}\\[2pt]
\lbl{Sentiment \& Emotion}
\atag{tagneg}{tagnegtxt}{Negative} $\cdot$ \atag{tagemotions}{tagemotionstxt}{Fear}\\[2pt]
\lbl{Topic} Gun policy \;
\lbl{Media Frame}
\atag{tagcrime}{tagcrimetxt}{Crime} \atag{tagcrime}{tagcrimetxt}{Security}\\[2pt]
\lbl{Semantic Frame}
\atag{tagsemframe}{tagsemframetxt}{Preventing} [\textit{ban}; agent: federal govt.] $\cdot$
\atag{tagsemframe}{tagsemframetxt}{Killing} [\textit{cause deaths}; cause: govt. ban]\\[2pt]
\lbl{Opinion} holder: author $\cdot$ gun ban is ineffective and harmful\\[2pt]
\lbl{Argument} premise: criminals bypass bans $\cdot$ conclusion: gun ban kills innocents\\[2pt]
\lbl{Claim} \textit{``A firearm ban will only cause deaths of innocent citizens.''} [central statement]
\end{minipage}
\end{tcolorbox}
\caption{Examples illustrating all perspective-related concepts across two texts.}
\label{fig:examples}
\end{figure*}

\subsection{Topics}
\label{sec:topics}

The choice of which topics to present contributes to perspective, because it is inherently linked to \textbf{selection} and framing bias \cite{rodrigo2024systematic}. Besides this, topic modeling has been used for perspective detection, leveraging unsupervised methods such as LDA to uncover latent \changed{\textit{semantic}} structures \cite{lin2008joint, ahmed2010staying, nguyen2013lexical, tsur2015frame, ajjour2019modeling}. Here, a perspective is seen as an aggregation of documents or text segments that share topical distributions. This approach is conceptually related to argument clustering (cf. §\ref{sec:claims}), but relies primarily on \changed{\textit{lexical}} patterns rather than argumentative structures \changed{and fine-grained \textit{discourse}} signals.

\paragraph{Perspective Signals} The ideological dimension is obtained from the topic representation as a latent variable: texts are assigned one weight for their topic and another one for their ideology, so that the latter can be isolated. This process is used to find political ideologies \cite{lin2008joint, ahmed2010staying, vilares2017detecting, nguyen2013lexical, roberts2014structural} and frames \cite{dimaggio2013exploiting, tsur2015frame, ajjour2019modeling} and can also enhance opinion mining \cite{draws2020helping}. Despite their decent performance, topic models risk oversimplifying perspective by focusing too heavily on \changed{\textit{lexical}} distributions rather than higher-level linguistic features. The alternative is combining them with \changed{\textit{semantic }}and \changed{\textit{discourse}} features to capture viewpoints more holistically \cite{carlebach2020news} (cf. §\ref{sec:claims}).

\subsection{Interactions Among Concepts}
\label{sec:interactions}

This section summarizes how the concepts can be combined for perspective detection and
analysis\changed{, as
a theoretically motivated account open to future empirical
investigation.} \changed{Figure~\ref{fig:examples} provides two
examples annotated with all the discussed concepts.}

Political ideology is the prominent conceptualization of \textit{perspective} in NLP \cite{pang2008opinion}. It is an overarching belief system \changed{rooted in a shared system of values that aggregates multiple} fine-grained viewpoints on entities, topics, and issues (§\ref{sec:ideology}). For example, a \textsc{conservative} ideology will be the sum of specific positions on various topics (e.g., \textsc{Against} migration, \textsc{Against} public health, etc.). These positions, intended as stances, sentiments or opinions, can serve as proxies for ideology detection \cite{grefenstette2004coupling, lin2006side, van-son-etal-2014-hope, bhatia2018topic, zhang-etal-2022-kcd}.
In contrast to ideology, these concepts are tied to  specific situations and are  variable across time and topics; e.g., a structured opinion will have a specific time, location, holder, and be linked to a specific event (cf. §\ref{sec:opinion}). Such fine-grained beliefs are encoded in argumentation, making this dimension the core nucleus of perspective and a concrete level where perspective can be identified. For example, to fully understand why someone has certain thoughts towards migration (opinion), a certain position towards the topic (stance), and a certain affective attitude (sentiment), we must consider the reasons deeply encoded in argumentation. Accordingly, some studies propose to find perspectives by clustering similar arguments and corresponding opinions (cf. §\ref{sec:claims}). 

In parallel, the choice of \textit{what} information to present and \textit{how} to present it is a good proxy for perspective. Media frames present a text under a particular light, and semantic frames induce \textit{perspectivization}. Both can be combined with the abstract beliefs discussed above to represent perspectives  holistically, aggregating the ideological belief and their concrete realizations in language and cognition \cite{card2015media, alashri2015animates, mendelsohn-etal-2021-modeling, tsur2015frame, field2018framing, card2015media, morstatter2018identifying, draws2022comprehensive, blokker22}.

\begin{table*}[tb!]
\centering\small
\setlength{\heavyrulewidth}{0.05em}
\setlength{\lightrulewidth}{0.03em}
\begin{tabular*}{\textwidth}{@{\extracolsep{\fill}}p{2.6cm}ccccc p{2cm}}
\toprule
\textbf{Concept} & \textbf{Ling.\ cues} & \textbf{Granularity} & \textbf{Entity-spec.} & \textbf{Disc.\ classes} & \textbf{Cluster} & \\
\midrule
\changed{Values}  & \scorecell{1.33}{0.58} & \scorecell{1.00}{0.00} & \scorecell{1.00}{0.00} & \scorecell{3.00}{1.73} & \changed{1} & \multirow{5}{*}{\textit{Val. \& ideology}} \\
\changed{Morals}  & \scorecell{2.00}{1.00} & \scorecell{1.00}{0.00} & \scorecell{1.33}{0.58} & \scorecell{3.00}{1.73} & \changed{1} & \\
Political ideology  & \scorecell{3.00}{1.00} & \scorecell{1.67}{0.58} & \scorecell{3.00}{2.00} & \scorecell{2.00}{1.73} & 1 &\\
Ideology bias       & \scorecell{1.67}{0.58} & \scorecell{1.00}{0.00} & \scorecell{2.00}{1.73} & \scorecell{1.00}{0.00} & 1 & \\
Political leaning   & \scorecell{2.33}{0.58} & \scorecell{1.00}{0.00} & \scorecell{2.33}{1.53} & \scorecell{1.33}{0.58} & 1 & \\
\midrule
Sentiment           & \scorecell{3.33}{0.58} & \scorecell{3.33}{1.15} & \scorecell{3.33}{0.58} & \scorecell{1.33}{0.58} & 2 & \multirow{4}{*}{\textit{Sent.\ \& stances}} \\
Polarity            & \scorecell{3.67}{0.58} & \scorecell{3.33}{1.15} & \scorecell{3.33}{1.15} & \scorecell{1.00}{0.00} & 2 & \\
Stances             & \scorecell{2.67}{1.53} & \scorecell{3.00}{2.00} & \scorecell{3.67}{1.53} & \scorecell{1.00}{0.00} & 2 & \\
\changed{Emotions} & \scorecell{3.33}{0.58} & \scorecell{3.00}{1.73} & \scorecell{3.67}{1.53} & \scorecell{2.67}{0.58} & \changed{2} & \\
\midrule
Media frames        & \scorecell{2.67}{0.58} & \scorecell{3.00}{1.00} & \scorecell{2.00}{1.00} & \scorecell{3.67}{0.58} & 3 & \multirow{2}{*}{\textit{Topics \& MF}} \\
Topics              & \scorecell{3.67}{0.58} & \scorecell{2.00}{1.00} & \scorecell{2.33}{1.53} & \scorecell{4.67}{0.58} & 3 & \\
\midrule
Arguments           & \scorecell{2.33}{1.15} & \scorecell{4.33}{0.58} & \scorecell{4.00}{0.00} & \scorecell{5.00}{0.00} & 4 & \multirow{4}{*}{\textit{Argumentation}} \\
Opinions            & \scorecell{2.00}{1.00} & \scorecell{4.00}{0.00} & \scorecell{4.00}{0.00} & \scorecell{5.00}{0.00} & 4 & \\
Claims              & \scorecell{3.67}{2.31} & \scorecell{5.00}{0.00} & \scorecell{4.33}{0.58} & \scorecell{5.00}{0.00} & 4 & \\
Semantic frames     & \scorecell{4.67}{0.58} & \scorecell{5.00}{0.00} & \scorecell{3.00}{2.00} & \scorecell{4.67}{0.58} & 4 & \\
\midrule
Average IRR ($\rho$) & 0.31 & 0.77 & 0.26 & 0.80 & & \\
\bottomrule
\end{tabular*}
\caption{Per-concept scores for four properties: (i) strength of linguistic cues, (ii) granularity, (iii) entity-specificity, and (iv) number of discrete classes. \changed{Mean ${\pm}$ std.\ dev.\ across annotators is reported (higher=red, lower=green). IRR: inter-rater reliability (Spearman's $\rho$).}}
\label{tab:annotation}
\end{table*}

\section{Structuring the Space of Perspective-related Concepts}
\label{sec:conceptualization}

In this section, we carry out \changed{an analysis} to investigate the organization of the perspective-related concepts (\textbf{RQ2}). 
We aim at inducing a property-driven structure over these concepts from expert judgments.
To do so, we first manually annotate them with a set of properties, then cluster the resulting distributions by similarity, and finally construct a hierarchy based on the clusters. 

\subsection{Annotating Conceptual Properties}
\label{sec:annotating}


We first annotate each concept along four properties that we characterize during the literature review. The definitional properties discussed in
§\ref{sec:perspectives}, such as \textbf{polar} and \textbf{affective}, summarized in Table~\ref{tab:properties_sent} for opinion-related terms,
and later formalized in the decision tree (Figure \ref{fig:diagram}), 
are binary and
concept-specific: they apply only to subsets of concepts and are not
gradable, making them unsuitable for a comparison across
all concepts. \changed{Therefore, we inductively derive four additional dimensions from the literature in §\ref{sec:perspectives} that are both
\textit{universal} (applicable to every concept in the survey) and
\textit{gradient} (varying continuously across concepts).} 
\changed{These properties are: (i) \textit{strength of linguistic
cues}: how strongly the concept is associated with specific linguistic
elements; (ii) \textit{granularity (scope)}: the typical scope or
localization of the concept within a text; (iii)
\textit{entity-specificity}: how strongly the concept is tied to
specific entities; (iv) \textit{number of discrete classes}: in classification, how many classes are used.} 
This \changed{property annotation} is performed independently by the three authors in their capacity as experts for the literature discussed above. We acknowledge the limitations of having 3 annotators only, but we judge it to be the best choice given the gained familiarity with the literature, and therefore, the required expertise to perform the type of annotations reliably. Based on \changed{a codebook (cf. Appendix~\ref{app:guidelines})}, they locate concepts on a Likert scale from 1 to 5 with respect to each property.



\paragraph{Results} \changed{Table \ref{tab:annotation}
reports the resulting score averages over the three annotators. The colors indicate the standard deviation (higher=red, lower=green). We report average Spearman $\rho$ for each property as a measure of inter-rater reliability (IRR).} 

\changed{The property with the highest IRR is \textit{number of discrete classes} ($\rho$ = 0.80), followed by \textit{granularity} ($\rho$ = 0.77). \textit{Strength of linguistic cues} and \textit{entity-specificity} have significantly lower agreement ($\rho$ = 0.31 and 0.26). Arguments and claims are difficult to agree on in \textit{strength of linguistic cues} because they are not systematically associated with certain recurrent linguistic cues, but at the same time they are identified with the linguistic expression itself. Stances in turn can also be ambiguous because they can be interpreted either as the ideological position toward an issue or as its concrete instantiation (similar to opinions). \textit{Entity-specificity} raises problems for concepts that can optionally refer to specific entities or not; for example, semantic frames involve entities and their roles, but they may or may not be necessary for annotation and detection. In sum, we take our result to be sufficiently robust for a clustering analysis.}


\subsection{Clustering Perspective Concepts} 
\label{sec:clustering}
\begin{figure}[tb!]
  \centering  \includegraphics[width=1\linewidth]{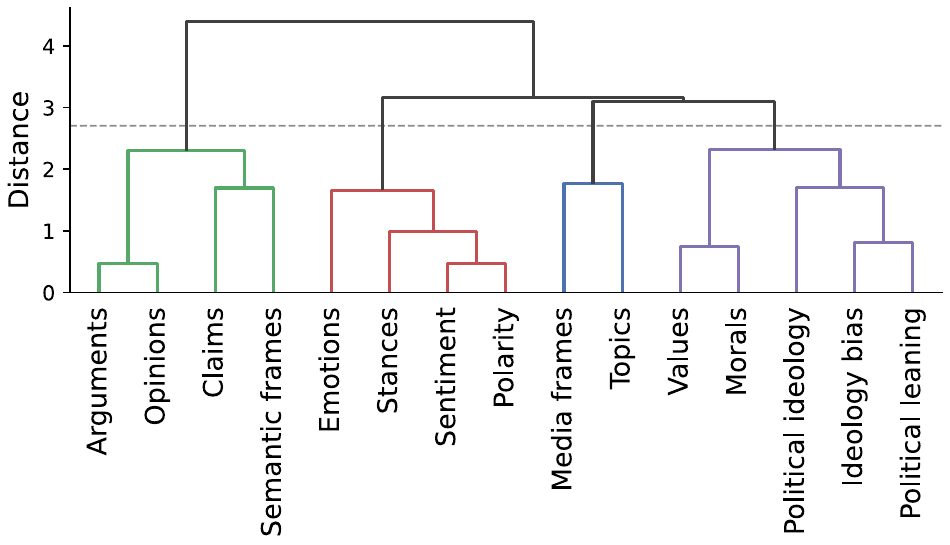}
  \caption{Dendrogram of concepts clusters obtained with hierarchical clustering. \changed{The dashed line marks the four-cluster cut.}}
  \label{fig:dendrogram}
\end{figure}

To group the concepts in Table~\ref{tab:annotation},
we apply agglomerative hierarchical clustering on annotator-aggregated scores, specifically \changed{average linkage} with Euclidean distance \citep{nielsen2016hierarchical}. \changed{Average linkage defines the distance
between two clusters as the mean of all pairwise distances between
their members, and iteratively merges the closest pair}. This results in \changed{four} compact groups, as shown in Figure \ref{fig:dendrogram}.

We refer to them as \changed{follows:} \changed{\textit{values and ideology}, the most abstract and global level, comprising overarching belief systems (§\ref{sec:values}) and political ideologies (§\ref{sec:ideology});} 
\textit{sentiment and stances}, including specific beliefs that express personal positions (§\ref{sec:sentiment}, §\ref{sec:stance}); \textit{topics and media frames}, \changed{including topical dimensions (§\ref{sec:frames}, §\ref{sec:topics});} \textit{argumentation}, including arguments and claims (§\ref{sec:claims}), semantic frames (§\ref{sec:frames}), and opinions (§\ref{sec:opinion}), all identifiable with linguistic structures. 

\begin{figure}[tb!]
  \centering  \includegraphics[width=1\linewidth]{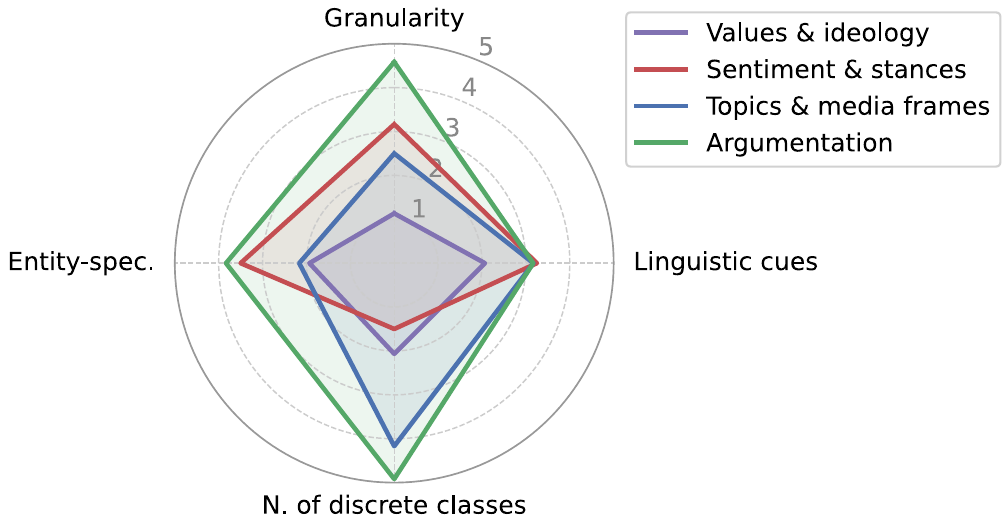}  \caption{Average scores for each conceptual cluster on the four properties.}  \label{fig:properties}
\end{figure}

The spider chart in Figure \ref{fig:properties}
shows the average scores for the  clusters on each property. \changed{The chart demonstrates the properties we consider are strongly correlated at the cluster level: the ranking of the clusters regarding the different properties agrees almost perfectly.
The only exception is that the \textit{sentiment and stances} cluster shows a lower class count than expected.
This is presumably the case because the cluster includes concepts whose interpretation varies depending on their theoretical definition and operationalization, notably emotion \citep{scarantino2016}. However, we consider this variation is minor for our ordering purposes.%
}

\subsection{A Model of Perspective}
\label{sec:hierarchy}

Given the strong correlation between the four properties which we see in Figure \ref{fig:properties}, \changed{we investigate whether the perspective-related concepts can be reduced to a single axis by running a Principal Component Analysis (PCA) of the annotations (cf. results in Appendix \ref{app:pca}). We find that this is largely the case: the first principal component (PC1) explains 62\% of the variance and shows a positive loading with each of the properties. When we represent all concepts purely in terms of their value on the dimension formed by PC1, we recover the four clusters almost perfectly, with the only outlier a swap between topics and stances (cf. Figure \ref{fig:pc1}). This likely happens because the two concepts received mid-point scores for all properties, apart from the class number: when giving more importance to such property, they get pulled apart (cf. PC2 in Figure \ref{fig:pc1_2}).} 

This result supports our interpretation that there is a latent linear ordering underlying the concepts. The axis identified by PC1 \changed{captures both linguistic and conceptual aspects, and} can be interpreted as a dimension of (generic) \textit{specificity}. Figure~\ref{fig:typo} is informed by this analysis and shows our model: the space is represented as a set of concentric circles,
ranging from ideological beliefs (outer) to linguistic instantiations (inner). At one end, we find generic concepts, such as ideology, which are not bound to specific situations, but underlie other fine-grained perspectives; they emerge throughout the document mainly with lexical cues, and map onto a few labels. At the other end, we find argumentation-related concepts, which are instead more specific both in what they express, i.e., precise situations and entities, and in how they are expressed: well localized in language spans, signalled by semantic and syntactic patterns, and mapping to a wide or open-ended class range.

\paragraph{Additional Factors}
So far, we have not considered information about the writer, annotator, or media source, even though they are indicators of perspective \cite{frenda2024perspectivist}. 
We include them in a separate box, since they describe the (extralinguistic) \textit{context} of the text rather than its \textit{content}. Indeed, metadata may not match the perspectives expressed in the text \cite{baly-etal-2020-detect}, and it is important to distinguish between grouping emerging from texts and groupings based on external data \cite{vitsakis2024voices}. We identify three types of extra-textual factors, based on the perspective holder: (i) the author's characteristics, (ii) the annotator’s characteristics and (iii) the media source. These include socio-demographic, political and cultural background (e.g., political orientation, gender, country, social affiliations, editorial stance), \changed{as well as annotation-related information (e.g., IAA)}. 

\paragraph{Previous Hierarchies}

Some other works in NLP aim at organizing the conceptual space of perspective. \citet{doan2022survey} divide the methods for perspective detection into: (i) political ideologies/leaning/party detection, (ii) political stance/framing detection, and (iii) political viewpoint extraction. \citet{klebanov2010vocabulary} distinguish four levels of perspective, from less to more abstract: (i) opinions, (ii) stances on specific issues, (iii) ideological positions, and (iv) demographic factors and life  of the author (e.g., place of birth, religion, culture, political tradition). Most similar to our proposal is the hierarchy by \citet{van-der-meer-2024-facilitating}, comprising three levels of abstraction: (i) stances, (ii) arguments, and (iii) values. 

Our model makes a number of contributions: (i) we include subjectivity-related concepts that have traditionally been treated separately \cite{pang2008opinion}; (ii) we include the linguistic level, following the idea that perspectives can be identified through arguments \cite{van-der-meer-2024-facilitating} and semantic patterns \cite{minnema-etal-2022-sociofillmore}; 
(iii) we annotate conceptual properties, providing an empirical support for the hierarchy; (iv) we distinguish \changed{\textit{content}} from \changed{\textit{context}-related} factors.

\section{Discussion}
\label{sec:discussion}


The goal of our study was to clarify and structure the space of  concepts relevant for research on \textit{perspectives} in NLP. We ask two research questions: \textbf{RQ1}, what concepts are used in perspective identification. We identify 15 concepts and characterize both their definition and operationalization based on a literature analysis (§\ref{sec:perspectives}, cf. Table~\ref{tab:terminology}). In \textbf{RQ2}, we ask how these concepts are related. The analysis of our expert annotations of four properties 
found that perspective-related concepts can be organized along a single dimension of linguistic and conceptual specificity that captures most of the variance between the concepts (§\ref{sec:conceptualization}, cf. Figure~\ref{fig:typo}).

\subsection{Outlook: Actionability}
\begin{figure*}[t!]
  \centering
  \includegraphics[width=1\linewidth]{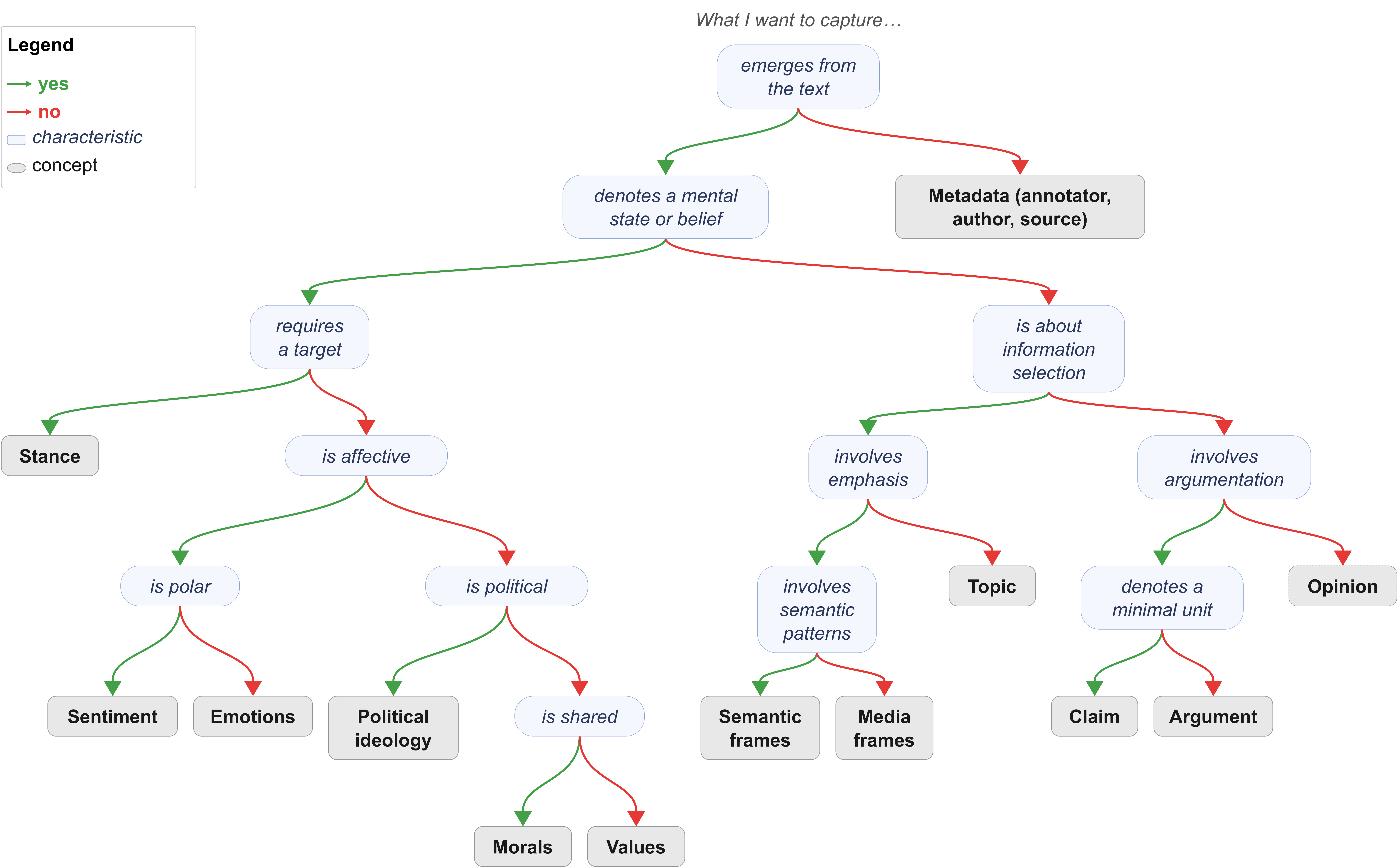}
    \caption{\changed{Decision tree for choosing what perspective concept(s) to adopt [is shared].
    \tikz[baseline=-0.5ex]{\draw[-{latex}, green!60!black, line width=2pt] (0,0) -- (0.55,0);} = \textit{yes},
    \tikz[baseline=-0.5ex]{\draw[-{latex}, red!90!black,   line width=2pt] (0,0) -- (0.55,0);} = \textit{no}. Discriminative characteristics are marked in \textbf{bold} in the literature review (§\ref{sec:perspectives})}}
  \label{fig:diagram}
\end{figure*}

Figure \ref{fig:typo} orders them in terms of specificity but does not provide guidance for choosing which one(s) to use in a hypothetical application. In Figure \ref{fig:diagram}, we  present a decision tree that leverages the features from §\ref{sec:perspectives} to guide concept selection. The following scenarios illustrate how the tree can be used. 

\paragraph{Scenario 1} I am studying a corpus of news that is fully topic- and issue-agnostic. I want to detect political perspectives for building a diverse news recommender. Following the decision tree, I decide to consider perspectives that emerge from the text. I aim to capture generic beliefs (\textit{emerges from the text} > \textit{denotes a mental state or belief}), without identifying explicit targets or relying on affective features, as the dataset is unstructured and mostly comprises factual news. The proposed operationalization is political ideologies. If I want to derive perspectives bottom-up from concrete language patterns (... > \textit{denotes a mental state or belief} = NO > \textit{is about information selection} = NO), I could start from opinion mining: as discussed, perspectives can be inferred by aggregating minimal positions on smaller topics (§\ref{sec:claims}). For diversification purposes, these opinions should then be reduced to a small number of meaningful clusters or categories.

\paragraph{Scenario 2} I am analyzing a corpus of news articles from different outlets covering the same event, with the goal of comparing how it is presented across sources. In this case, I have more flexibility, as I am not constrained by a fixed topic or application, and I do not need to cluster articles. My focus is on \textit{how} information is organized and empathized rather than on \textit{what} content is conveyed. Following the tree (... > \textit{is about information selection} > \textit{involves emphasis}), the proposed device is frames. In case I care about linguistic patterns and event structures, I could work with semantic frames. 

\paragraph{Scenario 3} I am building a politically-aligned LLM-based persona. I could leverage metadata about authors’ demographics from a corpus to guide the alignment (\textit{emerges from the text} = NO). Otherwise, the persona can be aligned with a broader political leaning which emerges from a consistent pattern of opinions across multiple issues. If my analysis is more granular, I could control for stances toward specific issues or targets (... > \textit{requires a target}) (e.g., \textsc{Pro} migration, \textsc{Pro} same-sex marriage, \textsc{Against} gun control). If I care about the affective tone (... > \textit{is affective}), I may consider sentiment or emotions as complementary dimensions.

\subsection{Future Research Directions}

Newspapers make editorial decisions at multiple levels of perspective,
including how to frame events, which arguments to use, what topics to
cover, and which stances to adopt. \changed{Despite lacking such deliberate
mechanisms,} LLMs convey perspectives emerging from training data in a comparable way. These viewpoints, embedded in textual choices, often
go unnoticed by readers. We claim that detecting, controlling, and
communicating these layers with transparency is worth-while to
support people's access to information and promote critical engagement.

\changed{By surveying perspective concepts, we have shown how analyzing argument structures jointly with information selection and presentation can map specific opinions to beliefs at different levels of granularity, up to ideology and values, while framing can reveal hidden \textit{over-emphasizing} signals. Exploring how these levels can be integrated into a coherent representation is a promising research direction in support of critical social analysis. Possible outcomes of this paper include a comprehensive annotation scheme, a modeling recipe, or an evaluation protocol for perspectives in text.}

On a more operational level, our conceptual hierarchy also carries direct implications for how we evaluate and audit language models. Rather than treating perspective bias as a monolithic property, the specificity axis offers a diagnostic lens: bias in LLMs may manifest differently at different levels, from systematic skews in ideological framing that pervade entire outputs, to more localized choices in argumentation structure or semantic framing that subtly shift responsibility or salience. Benchmarks for perspective diversity in generated text could be designed to probe each level independently, yielding a richer picture of where training data or alignment procedures introduce distortions. 

At the same time, the normative framing underlying much of this work -- that greater perspective diversity is inherently desirable -- deserves scrutiny. Diversity of perspectives is a meaningful democratic value when it reflects the genuine range of informed viewpoints on a contested issue; it becomes problematic when operationalized in ways that treat fringe or harmful positions as simply another point on a spectrum to be represented. Our hierarchy may help draw this distinction more precisely: diversity at the level of values and ideology calls for different normative criteria than diversity at the level of claims or arguments, where factual accuracy and logical coherence impose additional constraints beyond mere representational balance. Navigating this tension -- between pluralism and epistemic responsibility -- is as important as the development of methods to evaluate generated text. 




\bibliography{anthology,custom}
\bibliographystyle{acl_natbib}

\appendix




\section{Paper Collection}

\subsection{ACL regular expression search}
\label{app:acl_search}

The RegEx search is conducted on ACL titles and abstracts in May 2025,
using the following query: \textsc{perspective[s]} OR
\textsc{viewpoint[s]} AND \textsc{diversity} OR \textsc{news}. We
filter out papers where (i) \textit{perspective} is used generically
(e.g., to denote research angles or methods); (ii) the notion of
perspective is not central to the paper's contribution; or (iii) the
focus is on perspective-taking in a narrative sense (e.g., deictic
shifts), retaining 60 papers from the original 139 retrieved.

\subsection{Collected references}
\label{app:papers}
\changed{The references in Table \ref{tab:all} include
both papers from the ACL search and additional foundational work
collected manually via citation chaining, as described in
§\ref{sec:collection}.} We report all consulted
references, organized by thematic area.

\begin{table*}[t!]
\footnotesize\centering
\setlength{\tabcolsep}{4pt}
\renewcommand{\arraystretch}{0.88}
\begin{tabular}{p{2.8cm}p{13cm}}
\toprule
\textbf{Domain} & \textbf{References}\\
\midrule
Grounding work \& surveys & discourse/narrative: \citet{sanders1993linguistic}, \citet{graumann2008perspective}, \citet{van-der-meer-2024-facilitating}, \citet{van2025discourse}; ideology: \citet{van1998ideology}; bias: \citet{rodrigo2024systematic}; subjectivity: \citet{Quirk1985G}; sentiment: \citet{pang2008opinion}, \citet{pang2002thumbs}, \citet{munezero2014they}, \citet{liu2012survey}, \citet{chaturvedi2018distinguishing}; emotions: \citet{ekman1992argument}, \citet{plutchik1980general}; stance: \citet{kuccuk2020stance}, \citet{kiesling2018interactional}; political ideology: \citet{doan2022survey}; frames: \citet{entman1993framing}, \citet{semetko2000framing}, \citet{ali2022survey}, \citet{fillmore1976frame}, \citet{baker1998berkeley}, \citet{rashkin2016connotation}, \citet{otmakhova-etal-2024-media}, \citet{otmakhova-frermann-2025-narrative}; perspectivism: \citet{basile2022bias}, \citet{frenda2024perspectivist}, \citet{valette2024does}, \citet{mieleszczenko2023capturing}, \citet{aroyo2015truth}, \citet{muscato-etal-2024-overview}, \citet{sarumi-etal-2024-corpus}, \citet{van-der-meer-etal-2024-annotator}, \citet{frenda-etal-2023-epic}; claims/args: \citet{govier2010practical}, \citet{lauscher2022scientia}, \citet{cabrio2018five}; morals/values: \citet{schwartz1992universals}, \citet{graham2009liberals}, \citet{inglehart2000world}, \citet{atkinson2021value}, \citet{d_f_2026}, \citet{vida-etal-2023-values}; evaluative\ language: \citet{hunston2000evaluation}, \citet{benamara2017evaluative}; topics: \citet{alghamdi2015survey}, \citet{zhao2021topic}\\
Communication studies & conceptual: \citet{loecherbach2020unified}, \citet{mcquail1993media}, \citet{napoli1999deconstructing}, \citet{sunstein2007republic}, \citet{mcdonald2003conceptualization}, \citet{snow2000clarifying}, \citet{benson2009makes}, \citet{baden2017conceptualizing}, \citet{helberger2018exposure}, \citet{helberger2021democratic}, \citet{pariser2011filter}; applications: \citet{masini2017actor}, \citet{mulder2021operationalizing}, \citet{draws2021assessing}, \citet{vrijenhoek2021recommenders}, \citet{vossen2022creating}, \citet{draws2022comprehensive}, \citet{sorokovikova2024echo}\\
Morals \& values & \citet{sorensen2024value}, \citet{rottger2024political}, \citet{benkler2023assessing}, \citet{abdulhai2024moral}, \citet{roy2021identifying}, \citet{kiesel-etal-2022-identifying}\\
Private states & subjectivity: \citet{carbonell1979subjective}, \citet{riloff2003learning}, \citet{wiebe2004learning}, \citet{wiebe2005annotating}, \citet{shokri-etal-2024-subjectivity}, \citet{wilson-2008-annotating}; polarity/stance: \citet{wilson2005recognizing}, \citet{wilson2005opinionfinder}, \citet{somasundaran-wiebe-2010-recognizing}, \citet{choi2014+}, \citet{bosnjak-karan-2019-data}, \citet{liu2005opinion}, \citet{deng2015mpqa}; sentiment: \citet{martinc-etal-2021-embeddia}, \citet{yu2003towards}, \citet{ronningstad2024entity}, \citet{liu2010sentiment}; emotions: \citet{mohammad2013nrc}; opinion: \citet{kim2004determining}, \citet{grefenstette2004coupling}, \citet{morante2020annotating}, \citet{van2016grasp}\\
Media bias & \citet{yano2010shedding}, \citet{al2012automatic}, \citet{recasens2013linguistic}, \citet{fan2019plain}, \citet{vargas2023predicting}, \citet{mastrine2022spot}, \citet{lin2025investigating}, \citet{da2023overview}, \citet{saleh-etal-2019-team}, \citet{van-son-etal-2014-hope}\\
Polar ideology \& stance & \citet{lin2006side}, \citet{lin2008joint}, \citet{ahmed2010staying}, \citet{hardisty2010modeling}, \citet{klebanov2010vocabulary}, \citet{hasan2012predicting}, \citet{elfardy2015ideological}, \citet{johnson2016all}, \citet{zhao2024zerostance}, \citet{reuver-etal-2024-investigating}, \citet{roy2023tale}, \citet{anand2011cats}, \citet{mohammad2016semeval}\\
Political leaning & \citet{gentzkow2019measuring}, \citet{monroe2008fightin}, \citet{budak2016fair}, \citet{gangula-etal-2019-detecting}, \citet{baly-etal-2020-detect}, \citet{feng2021kgap}, \citet{zhang-etal-2022-kcd}, \citet{alzhrani2022political}, \citet{kim2023multi}, \citet{han-etal-2019-fallacy}, \citet{li-goldwasser-2021-mean}, \citet{kiesel2019semeval}, \citet{li-goldwasser-2019-encoding}, \citet{ceron2022optimizing}, \citet{martinez-etal-2024-balancing}, \citet{laver2003extracting}, \citet{slapin2008scaling}, \citet{jakob-etal-2024-augmented}, \citet{huguet-cabot-etal-2020-pragmatics}, \citet{bhatia2018topic}, \citet{webson-etal-2020-undocumented}, \citet{greene2009more}, \citet{iyyer2014political}\\
Frames & media frames: \citet{baumer2015testing}, \citet{lecheler2015effects}, \citet{alashri2015animates}, \citet{card2015media}, \citet{card-etal-2016-analyzing}, \citet{naderi2017classifying}, \citet{field2018framing}, \citet{morstatter2018identifying}, \citet{khanehzar2019modeling}, \citet{mendelsohn-etal-2021-modeling}, \citet{khanehzar2021framing}, \citet{gilardi2023chatgpt}, \citet{piskorski2023semeval}, \citet{bhatia-etal-2021-openframing}, \citet{das-etal-2024-media}, \citet{tourni-etal-2021-detecting-frames}, \citet{liu-etal-2019-detecting}, \citet{nakov-etal-2021-second}, \citet{daffara2025generalizability}, \citet{kwak2020systematic}, \citet{ziems-yang-2021-protect-serve}, \citet{sengupta-etal-2024-analyzing}, \citet{blokker22}; semantic frames: \citet{te2020framing}, \citet{xia-etal-2021-lome}, \citet{minnema2021frame}, \citet{minnema-etal-2022-sociofillmore}, \citet{meluzzi2021responsibility}, \citet{pinelli2021gender}, \citet{zanchi2021numeri}, \citet{minnema-etal-2023-responsibility}, \citet{minnema-etal-2022-dead}\\
Clusters as perspectives & \citet{chen2017opinion}; sub-groups: \citet{dasigi2012genre}, \citet{abu2013identifying}, \citet{abu2012subgroup}; claims/args: \citet{bar2017stance}, \citet{chen2019seeing}, \citet{ajjour2019modeling}, \citet{trabelsi2014finding}, \citet{carlebach2020news}, \citet{boland2022beyond}, \citet{stab2017parsing}; LDA: \citet{boydstun2013identifying}, \citet{tsur2015frame}, \citet{nguyen2013lexical}, \citet{roberts2014structural}, \citet{paul2010two}, \citet{thonet2016vodum}, \citet{menini2016agreement}, \citet{vilares2017detecting}; arg. retrieval: \citet{saha-srihari-2024-turiya-perpectivearg2024}, \citet{kang-etal-2024-xfact}; other: \citet{vitsakis2024voices}, \citet{de2005news}, \citet{reimers-etal-2019-classification}, \citet{dimaggio2013exploiting}, \citet{draws2020helping}\\
Multi-view generation & \citet{chen2025open}, \citet{plepi2024perspective}; personas: \citet{hu-etal-2025-debate}, \citet{baltaji-etal-2024-conformity}; chatbots: \citet{hamad-etal-2024-asem}; diversity: \citet{hayati-etal-2024-far}, \citet{park-etal-2019-generating}; reasoning: \citet{yan-etal-2024-mirror}; table-to-text: \citet{zhao-etal-2023-loft}; multichannel: \citet{chan-etal-2020-multichannel}; writing: \citet{spangher-etal-2024-llms}; fake news: \citet{zheng-etal-2025-unveiling}, \citet{wan-etal-2024-dell}; reframing: \citet{sheng-etal-2023-learning}, \citet{chen-etal-2021-controlled-neural}; alignment: \citet{sorensen-2024-pluralistic}\\
Multi-view summarization & \citet{barker-etal-2016-whats}, \citet{deas-mckeown-2025-summarization}, \citet{liu-etal-2021-multioped}, \citet{liu-etal-2024-p3sum}, \citet{van-der-meer-etal-2024-empirical}, \citet{huang-etal-2023-examining}, \citet{olabisi-etal-2022-analyzing}, \citet{ye-etal-2024-globesumm}; fake news: \citet{chang-etal-2023-beyond}\\
News recommendation & \citet{reuver-etal-2021-human}, \citet{reuver-etal-2021-nlp}, \citet{milbauer-etal-2023-newssense}; stance: \citet{ruder-etal-2018-360deg}; clickbait: \citet{shi-etal-2023-multiview}\\
\bottomrule
\end{tabular}
\caption{All references consulted, organized by thematic area.}
\label{tab:all}
\end{table*}

\section{Property Annotation and Analysis}
\label{app:guidelines}

\subsection{Annotation Procedure}

The annotation was conducted in two iterations. In the first iteration,
the three annotators independently assigned scores to each concept on the
four dimensions described in the codebook below (§\ref{app:codebook}).
After completing the first round, annotators shared their scores and
discussed cases of disagreement, focusing on cases where the operationalization was ambiguous. In the second iteration,
annotators revised their scores in light of the discussion, without being
required to reach consensus: final scores reflect each coder's
independent expert judgment. Inter-rater reliability was computed on
each property using Spearman $\rho$ (cf. §\ref{sec:conceptualization}).

\subsection{Codebook}
\label{app:codebook}

\subsubsection*{Instructions}
For each perspective-related concept listed in
Table~\ref{tab:annotation}, independently assign a score from 1 to 5
on each of the four dimensions below. Scores reflect your expert
judgment based on the NLP literature. In case of doubt, consult the concept definitions and examples in
Table~\ref{tab:terminology} at the end of this section. Do
not discuss your scores with other annotators until all annotations in the
current iteration are complete.

\subsubsection*{Dimensions}

We characterize perspective-related concepts along four dimensions,
selected because they jointly capture the key aspects that distinguish
concepts in the literature and determine how they are operationalized
in annotation and detection tasks. 

For each concept, assign a score on a Likert
scale from 1 to 5 (1~=~low; 5~=~high):

\begin{enumerate}

\item \textit{Strength of linguistic cues}: how strongly the concept
is associated with specific linguistic elements [1~=~weakly associated;
5~=~strongly associated]. This dimension captures to what extent a
concept is signalled by identifiable surface features: some concepts
leave strong lexical and syntactic traces, while others require holistic
document-level inference with no reliable surface cues.

\item \textit{Granularity (scope)}: the typical scope or localization
of the concept within a text [1~=~very broad/document-level;
5~=~very narrow/phrase- or clause-level]. This dimension determines
the annotation unit: some concepts are inherently local, while others
are distributed across an entire document.

\item \textit{Entity-specificity}: how strongly the concept is tied to
a specific entity (e.g., politician, policy, event) [1~=~not tied to
a specific entity; 5~=~directly tied to a specific entity]. This
dimension distinguishes target-generic from target-specific concepts,
determining whether entity recognition is a prerequisite for annotation and detection.

\item \textit{Number of discrete classes}: in a typical classification
task, how many classes the concept comprises [1~=~few classes,
e.g., binary; 5~=~many classes]. This dimension reflects the complexity of the label space: binary or ternary concepts organize reality in a coarse-grained fashion, while open-ended concepts require finer distinctions.

\end{enumerate}

\subsubsection*{Decision Rules}
\begin{itemize}
    \item If a concept can be operationalized in multiple ways
    (e.g., sentiment as binary emotional polarity or as the perspective expression itself), refer to the examples in Table \ref{tab:annotation} to choose a specific interpretation.
    \item Score each concept independently; do not let your score on one dimension influence another.
\end{itemize}

\subsubsection*{Anchor Examples and Concept Definitions}
For reference, the following examples illustrate prototypical high and
low scores across dimensions:

\begin{itemize}
    \item \textbf{High across dimensions} (score~$\approx$~5):
    \textit{claims} --- triggered by specific linguistic patterns, tied to a specific proposition,
    open-ended label space.
    \item \textbf{Low across dimensions} (score~$\approx$~1):
    \textit{ideology bias} --- no reliable surface cue,
    document-level, no entity required, binary label.
\end{itemize}

Table~\ref{tab:terminology} provides definitions and examples for each
concept to be annotated. Consult it when the scope of a concept is
unclear before assigning scores.

\begin{table*}[h!]
\footnotesize\centering
\begin{tabular}{p{2.3cm}p{7.9cm}p{4.5cm}}
\toprule
\textbf{Term} & \textbf{Definition} & \textbf{Example (class/value)}\\
\midrule
\textbf{Argument}& A set of statements about a controversial topic, made up of premises and conclusions \cite{govier2010practical}. & \textit{Marijuana should not be legalized. That's because sustained use of marijuana worsens a person's memory, and nothing that adversely affects one's mental abilities should be legalized.} \cite{govier2010practical}\\
\textbf{Claim}& A component of an argument, either the premise or the conclusion \cite{govier2010practical} or the central assertion \cite{stab2017parsing, boland2022beyond}. & \textit{Animals should not be used for scientific or commercial testing.} \cite{chen2019seeing}\\
\textbf{Semantic frames} & In linguistics, structures of meaning consisting of semantic roles and lexical units \cite{fillmore1976frame}. & \textsc{abuse}, \textsc{rape}, \textsc{cause\_motion}, \textsc{use\_firearm}\\
\textbf{Media frames}& Topic-like dimensions that organize reality to shape understanding and promote specific views \cite{card2015media}. & \textsc{morality}, \textsc{public sentiment}, \textsc{cultural identity}\\
\textbf{Topic} & An interpretable semantic concept emerging from probabilistic word distributions \cite{alghamdi2015survey}. & \textsc{Economy}, \textsc{Politics}\\
\textbf{Opinion}& An idea or belief about a target that contributes to a viewpoint \cite{munezero2014they, thonet2016vodum}, typically found in subjective texts \cite{pang2008opinion}. & \textit{Mary said the dress is beautiful.} \cite{kim2004determining}\\
\textbf{Stance}& The evaluation of a target, which can align the author with or against others \cite{kuccuk2020stance}. & \textsc{in favour}--\textsc{against}\\
\textbf{Sentiment}& A lasting feeling or disposition toward something \cite{munezero2014they}. & \textsc{positive}--\textsc{negative}\\
\changed{\textbf{Emotions}} & \changed{Affective states specifying the type of subjective response \cite{plutchik1980general}.} & \changed{\textsc{fear}, \textsc{anger}, \textsc{joy}, \textsc{sadness}}\\
\textbf{Polarity} & The orientation of sentiment towards positive or negative \cite{pang2008opinion}. & \textsc{positive}--\textsc{negative}\\
\textbf{Ideological bias} & It occurs when a text is unbalanced towards a particular ideology \cite{yano2010shedding}.& \textsc{biased}--\textsc{not biased}\\
\textbf{Political ideology} & The political orientation of a text in terms of ideological values \cite{pang2008opinion}. & \textsc{Palestinian}/\textsc{Israeli}, \textsc{conservative}--\textsc{liberal}\\
\textbf{Political leaning} & The political orientation of a text in terms of political spectrum \cite{doan2022survey}. & \textsc{left}--\textsc{right}\\
\changed{\textbf{Morals}} & \changed{Societal-level norms used to judge between ``right'' and ``wrong'' \cite{vida-etal-2023-values}.} & \changed{\textsc{Care/Harm}, \textsc{Fairness/Betrayal}}\\
\changed{\textbf{Values}} & \changed{The individual ideals that people
pursue \cite{sorensen2024value}} & \changed{\textsc{Freedom}, \textsc{Equality}}\\
\bottomrule
\end{tabular}
\caption{Perspective-related terms with definitions and examples. We report example labels if the concept is commonly mapped onto label sets for text classification, \changed{and example} values when it \changed{it is open-ended and} commonly denotes language spans in mining tasks.}
\label{tab:terminology}
\end{table*}

\subsection{Principal Component Analysis (PCA)}
\label{app:pca}

\begin{figure}[h!]
\centering  \includegraphics[width=\linewidth]{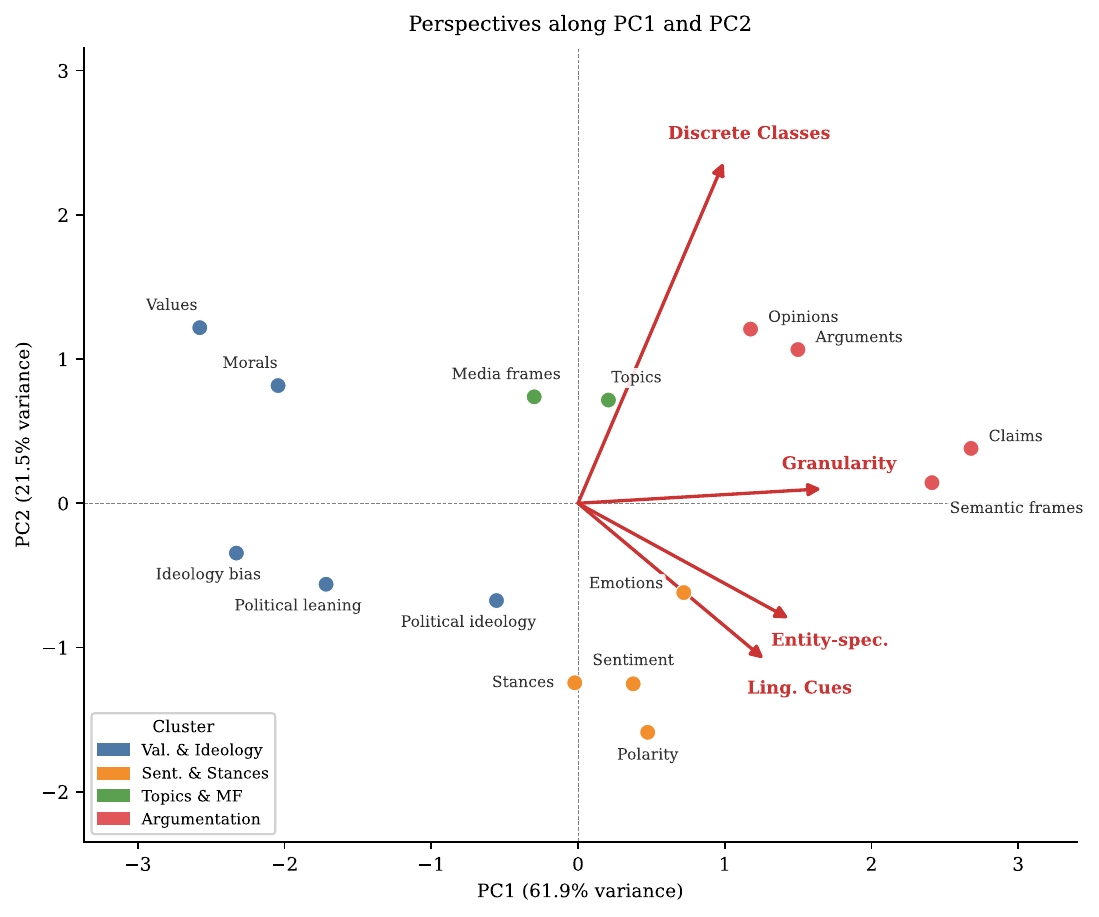}  \caption{\changed{Perspective concepts along PC1 and PC2. The structure validates the clusters found in §\ref{sec:clustering}.}}
\label{fig:pc1_2}
\end{figure}

\begin{figure}[tb!]
\centering  \includegraphics[width=1\linewidth]{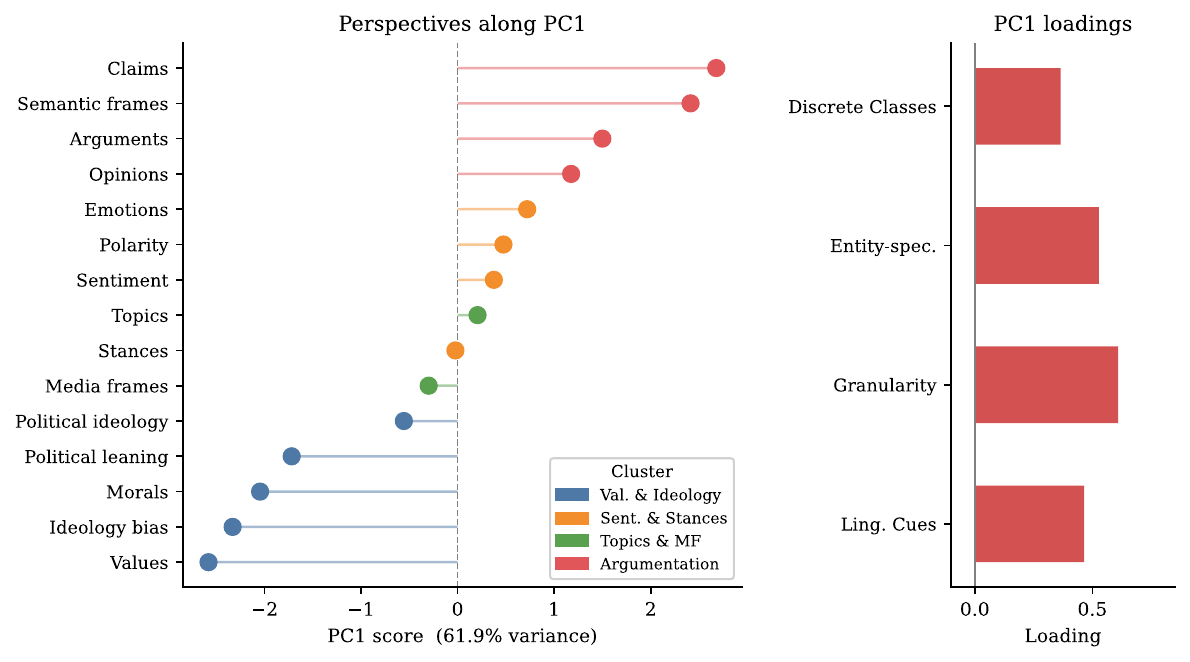}  \caption{\changed{Perspective concepts scores and property loadings on PC1. The four properties all positively correlate with PC1, validating our linear model in Figure \ref{fig:typo}.}}  \label{fig:pc1}
\end{figure}

We perform PCA on averaged annotations as a validation of our conceptual model shown in Figure \ref{fig:typo}. PC1 and PC2 explain 83.4\% of the variance. As shown in Figure \ref{fig:pc1_2}, the distribution of concepts along these two latent dimensions corresponds to the four clusters recognized in §\ref{sec:clustering}. Notably, PC1 alone explains 61.9\% of the variance and is positively correlated with each of the four properties (cf. Figure \ref{fig:pc1}). PC2 instead is dominated by the class number and, negatively, by the strength of linguistic cues. Overall, these findings validate the cluster analysis and the correlation between properties, supporting the linear organization of the concepts in clusters along a single axis of \textit{specificity}.

\end{document}